\documentclass[10pt,twocolumn,letterpaper]{article}

\usepackage[pagenumbers]{cvpr} % To force page numbers, e.g. for an arXiv version

\usepackage[accsupp]{axessibility}

\usepackage{graphicx}
\usepackage{amsmath}
\usepackage{amssymb}
\usepackage{booktabs}
\usepackage{adjustbox}
\usepackage[normalem]{ulem}
\usepackage{graphics}
\usepackage{array}
\usepackage{multirow}
\usepackage{wrapfig}
\usepackage{hhline}
\usepackage{pifont}% http://ctan.org/pkg/pifont
\newcommand{\showimagew}[2][\linewidth]{\includegraphics[width={#1}]{{#2}}}

\newcolumntype{K}[1]{>{\centering\arraybackslash}p{#1}}
\newcommand{\negvspace}{\vspace{-0.2cm}}
\usepackage{duckuments}
\usepackage{tikzducks}
\usepackage{graphicx}
\usepackage{adjustbox}
\usepackage{pifont}
\usepackage{float}

\usepackage{atbegshi}
\newcommand{\placetextbox}[3]{
  \setbox0=\hbox{#3}% Put <stuff> in a box
  \AtBeginShipoutNext{\AtBeginShipoutUpperLeft{%
    \put(\dimexpr#1\paperwidth\relax,-\dimexpr#2\paperheight\relax)
    {\vtop{{\null}\makebox[0pt][c]{#3}}}%
  }}%
}

\usepackage[pagebackref,breaklinks,colorlinks]{hyperref}

\usepackage[capitalize]{cleveref}
\crefname{section}{Sec.}{Secs.}
\Crefname{section}{Section}{Sections}
\Crefname{table}{Table}{Tables}
\crefname{table}{Tab.}{Tabs.}

\def\cvprPaperID{7402} % *** Enter the CVPR Paper ID here
\def\confName{CVPR}
\def\confYear{2023}

\begin{document}

\title{Computational Flash Photography through Intrinsics\negvspace\negvspace}

%\and

\author{
Sepideh Sarajian Maralan\quad\quad 
Chris Careaga\quad\quad
Ya\u{g}{\i}z Aksoy
\negvspace\\\\\negvspace
Simon Fraser University
}

\twocolumn[{%
\maketitle

\placetextbox{0.13}{0.02}{\includegraphics[width=4cm]{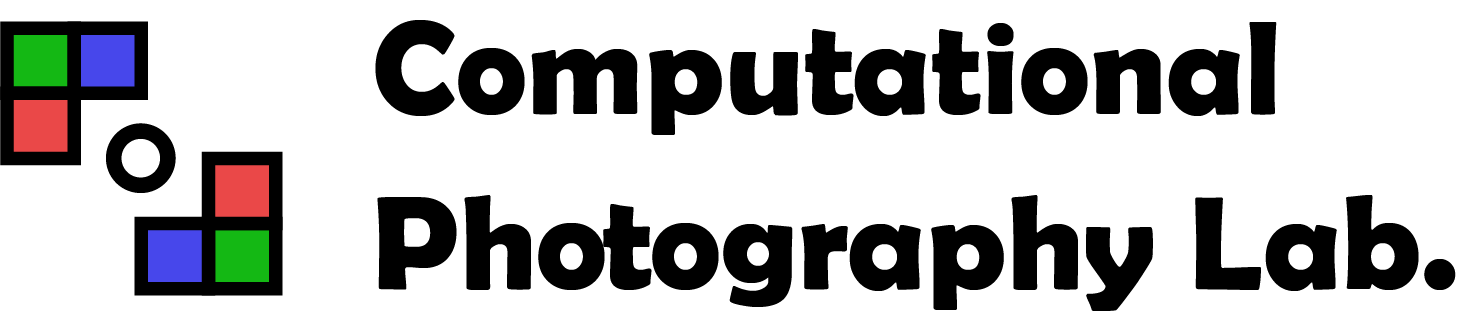}}
\placetextbox{0.78}{0.02}{Find the project web page here:}
\placetextbox{0.78}{0.035}{\textcolor{purple}{\url{http://yaksoy.github.io/intrinsicFlash/}}}

\begin{center}
    \centering
    \negvspace
    \negvspace
    \showimagew[\linewidth]{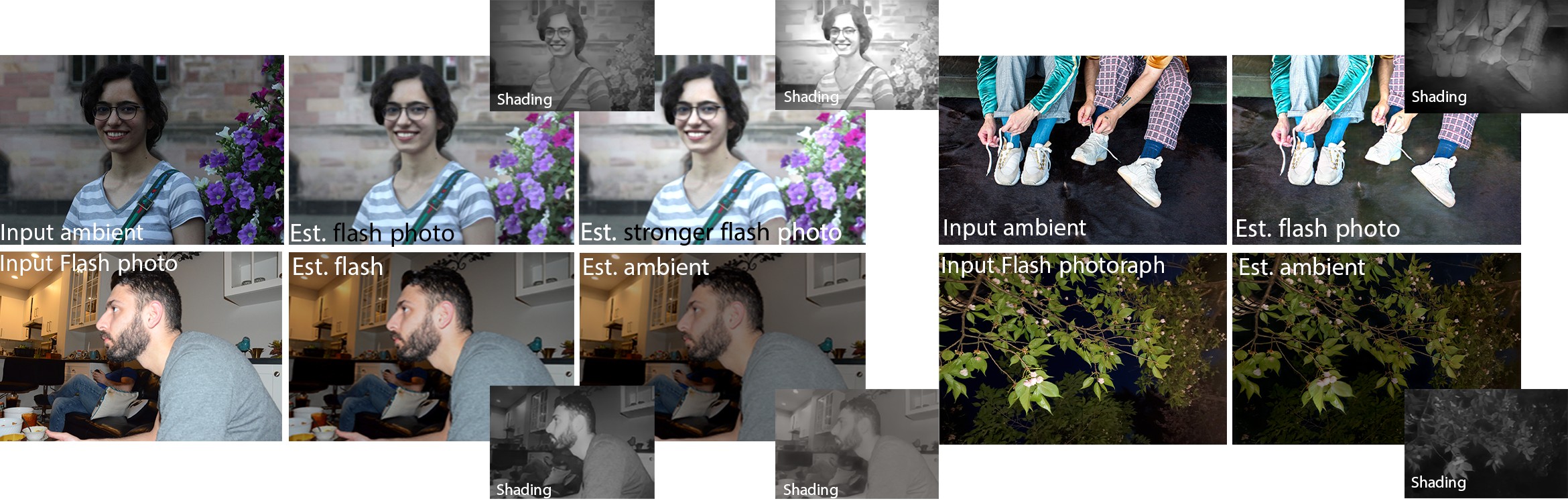}
    \captionof{figure}{
    We develop a system to computationally control the flash light in photographs originally taken with or without flash. 
    We formulate the flash photograph formation through image intrinsics, and estimate the flash shading through generation for no-flash photographs (top) or decomposition where we separate the flash from the ambient illumination for flash photographs (bottom).
    }
    \label{fig:teaser}
\end{center}%
}]

\newcommand{\chris}[1]{\textcolor{cyan}{{[Chris: #1]}}}
\newcommand{\sepideh}[1]{\textcolor{purple}{{[Sepideh: #1]}}}
\newcommand{\yagiz}[1]{\textcolor{blue}{{[Ya\u{g}{\i}z: #1]}}}

%%%%%%%%% ABSTRACT
\begin{abstract}
  
\negvspace
Flash is an essential tool as it often serves as the sole controllable light source in everyday photography. 
However, the use of flash is a binary decision at the time a photograph is captured with limited control over its characteristics such as strength or color. 
In this work, we study the computational control of the flash light in photographs taken with or without flash. 
We present a physically motivated intrinsic formulation for flash photograph formation and develop flash decomposition and generation methods for flash and no-flash photographs, respectively. 
We demonstrate that our intrinsic formulation outperforms alternatives in the literature and allows us to computationally control flash in in-the-wild images.

\negvspace
\negvspace
%The majority of common cameras have an integrated flash that improves lighting in a variety of situations, particularly in low-light environments. 
%Before capturing an image, the photographer must make a decision regarding the usage of flash. 
%However, flash strength cannot be adjusted once it has been utilised in an image.
%In this work, we target two application scenarios in computational flash photography: decomposition of a flash photograph into its illumination components and generating the flash illumination from a given single no-flash photograph. 
%We address these tasks through the intrinsic components which simplifies the tasks with the use of convolutional neural networks. 
%We present two distinct models for flash generation and decomposition to address subtasks unique to each problem, which were trained using a relatively large flash/no-flash dataset prepared by us.
\end{abstract}

%%%%%%%%% BODY TEXT

\section{Introduction}
\label{sec:intro}

\begin{figure*}
\centering
\includegraphics[width=\linewidth]{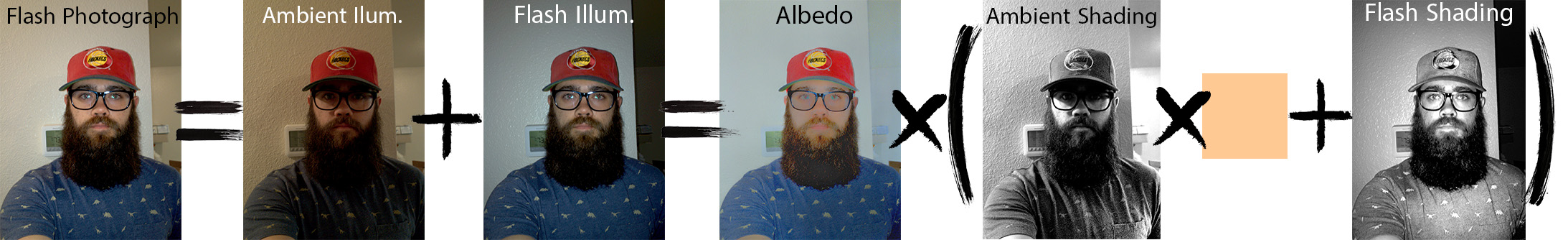}
\caption
{
We can model the flash photograph formation in terms of separate shading maps for flash and ambient lights with a shared albedo.
}
\negvspace\negvspace
\label{fig:formula}
\end{figure*}

Flash is an essential tool as it often serves as the sole controllable light source in everyday photography available on a wide range of cameras.
Although most cameras measure the environment light to adjust the flash strength, whether to use flash or not is a binary decision for the casual photographer without any control over flash characteristics such as strength or color temperature. 
In addition to hardware constraints, this limitation also stems from the typical use case of flash 
%illuminating the environment for a short instant 
in situations that do not allow lengthy experimentation with illumination. 
This makes the computational control of the flash a desirable application in photography.

Using the superposition of illumination principle, a flash photograph can be modeled as the summation of flash and ambient illuminations in linear RGB:
\begin{equation}
P = A + F,
\label{eq:superposition}
\end{equation}
where $P$, $A$, and $F$ represent the flash photograph, the ambient illumination, and the flash illumination respectively. 
To computationally control the flash post-capture, the ambient and flash illuminations need to be decomposed. 
Similarly, to \emph{add} flash light to a no-flash photograph, the flash illumination needs to be generated. 
In this work, we tackle both flash illumination decomposition and generation problems and show that regardless of the original photograph being captured with or without flash, we can generate and control the flash photograph computationally. 

We start by modeling the flash photograph formation through image intrinsics. 
Intrinsic image decomposition models the image in terms of reflectance and shading:
\begin{equation}
I = R \cdot S,
\label{eq:intrinsicmodel}
\end{equation}
where $I$, $R$, and $S$ represent the input image, the reflectance map, and the shading respectively. 
This representation separates the effects of lighting in the scene from the illumination-invariant surface properties. 
With lighting being the only difference between the ambient and flash illuminations, both the decomposition and generation problems are more conveniently modeled on shading. 
Assuming a mono-color ambient illumination and a single-channel shading representation, we re-write Eq.~\ref{eq:superposition} using Eq.~\ref{eq:intrinsicmodel}:
\begin{equation}
P = R \cdot c_A S_A + R \cdot S_F = R \cdot (c_A S_A + S_F),
\label{eq:intrinsicsuperposition}
\end{equation}
where $c_A$ is a 3-dimensional vector representing the color temperature of the ambient illumination assuming the flash photograph $P$ is white-balanced for the flash light. 
Fig.~\ref{fig:formula} visualizes our representation.

We develop a system for flash decomposition where we estimate $c_A$, $S_A$, and $S_F$ from a flash photograph and combine them with the shared reflectance $R$ to compute the final flash and ambient illumination estimations. 
We show that this physically-inspired modeling of the flash photograph allows us to generate high-quality decomposition results. 

Flash decomposition is an under-constrained problem where both $A$ and $F$ are partially observed within the input $P$. 
Flash generation, on the other hand, is a more challenging problem that amounts to full scene relighting as, in this scenario, only $A$ is given as input for estimating $F$. 
In addition to our intrinsic modeling, we formulate a self-supervision strategy for the generation problem through the cyclic relationship between decomposition and generation. 
We show that with a combined supervised and self-supervised strategy, we can successfully simulate flash photographs from no-flash input.

\negvspace
\section{Related Work}
\label{sec:related}

\begin{figure*}
\negvspace
\includegraphics[width=1\linewidth]{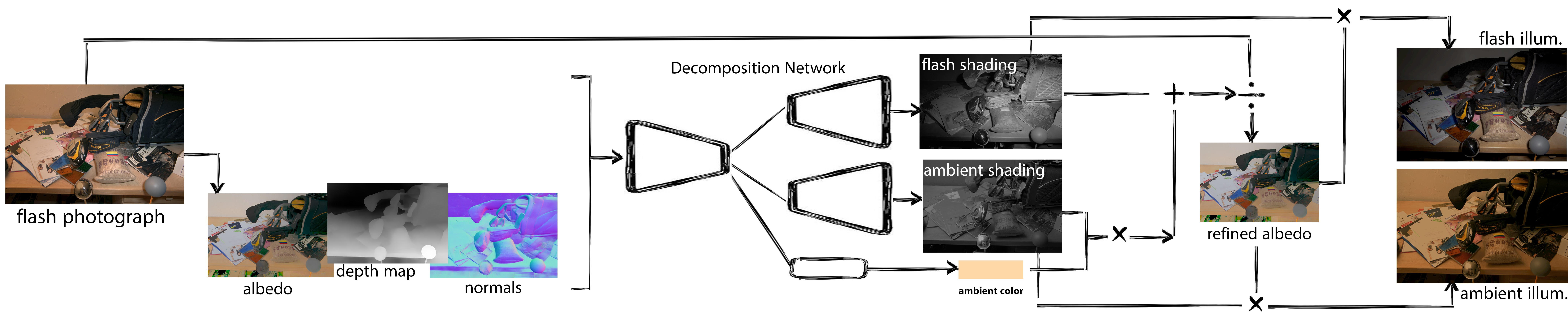}
\negvspace
\negvspace
\negvspace
\caption[Flash Shading Decomposition]{We provide the encoder with the flash photograph, as well as the albedo, surface normals, and depth map obtained through preprocessing. Two decoders and fully connected layers output the estimated flash shading, ambient shading, and ambient temperature. The ambient illumination is the element wise multiplication of the colored ambient shading with the refined albedo.
}
\negvspace
\negvspace
\label{fig:decompose}
\end{figure*}

Earlier work on flash photography mainly focused on using flash/no-flash pairs for image enhancement. 
The pair of photographs, although challenging to capture in real-world setups, have been shown to be useful for reasoning about the ambient light characteristics~\cite{Illum2001}, image enhancement and denoising~\cite{FIntrinsic2004}, artifact removal~\cite{Freflection2005,DigPhoto2004}, image matting~\cite{Fmatting2006}, white balancing~\cite{Fwhitebalance2016}, and saliency estimation~\cite{Fsaliency2014}. 

In recent years, multiple works \cite{flashambient, deepflash, multi2019} generated datasets of flash/no-flash image pairs. These datasets with innovative methods of image-to-image translation paved the way for the generation or decomposition of flash illumination. \cite{flashambient, deepflash} and, more recently, \cite{chavez2020ambient} tackled the problem of image relighting with flash illumination which is also the focus of our work. Aksoy et al.~\cite{flashambient} proposed a baseline network setup for flash illumination decomposition, Capece et al.~\cite{deepflash} presented a system to convert a flash photograph into a studio illumination, while Chavez et al.~\cite{chavez2020ambient} utilized a conditional adversarial network with attention maps to translate from flash to ambient image. Due to the correlation of flash light with scene depth, depth maps can be used to guide the networks for flash decomposition and subsequently for image relighting~\cite{qiu2020geometry}. 
In this work, we model the flash photograph formation through image intrinsics and show that our physically-inspired formulation outperforms the methods formulated as direct image-to-image translation.

Flash generation or decomposition can both be formulated as relighting problems. Image relighting is typically formulated with specific constraints such as portrait relighting or relighting of indoor/outdoor environments. These case-specific relighting methods often can not be directly applied to flash photography. For relighting of indoor and outdoor environments, Helou et al. present VIDIT~\cite{vidit}, a synthetic dataset with different varying directions and temperatures. Multiple works are trained using this dataset for one-to-one and any-to-any illumination transfer \cite{elhelou2020aim, helou2021ntire, DSRN, Yazdani2021PhysicallyID}. Zhu et al.~\cite{zhu2022ian} propose a specialized network that utilizes geometry information to perform relighting as a direct image-to-image problem. Similar to our approach, intrinsic decomposition components have been used as an intermediate representation in relighting approaches. Yazdani et al.~\cite{Yazdani2021PhysicallyID} construct the relit image by estimating the albedo and shading of the target. Wang et al.~\cite{wang2020deep} estimate a shadow-free image generated from multiple illuminations of a single scene. Both methods attempt to predict intrinsic components and relight simultaneously, which makes it a very difficult problem for networks with limited capacity. In our formulation, we utilize intrinsics from an off-the-shelf method and use the intrinsic formulation to simplify the relighting process for the network, allowing us to generate higher-quality results.

\section{Flash Illumination Decomposition}
\label{sec:decomposition}

We define the illumination decomposition task as estimating the separate flash and ambient illuminations given a flash photograph. 
In flash photography, most cameras auto-white balance according to the flash light, therefore the flash illumination appears white. 
Since most environments are under lighting that has a different color than flash, the ambient illumination appears with a color shift from the flash illumination. 
Assuming a mono-color ambient illumination, we model the flash photograph formation in the intrinsic domain using Eq.~\ref{eq:intrinsicsuperposition}, as visualized in Fig.~\ref{fig:formula}. 

Following our model, we re-frame the illumination decomposition problem as a shading prediction task. 
We propose a pipeline to generate shading components that correspond to the two illumination sources from a flash photograph. 
Our decomposition pipeline consists of a single encoder that feeds into three different decoders. 
Two decoders generate the single-channel ambient and flash shadings $S_A$  and $S_F$, while the third uses linear layers to regress the color temperature of the ambient illumination that defines $c_A$.
We present an overview of our pipeline in Fig.~\ref{fig:decompose}.

\begin{figure}
\includegraphics[width=1\linewidth]{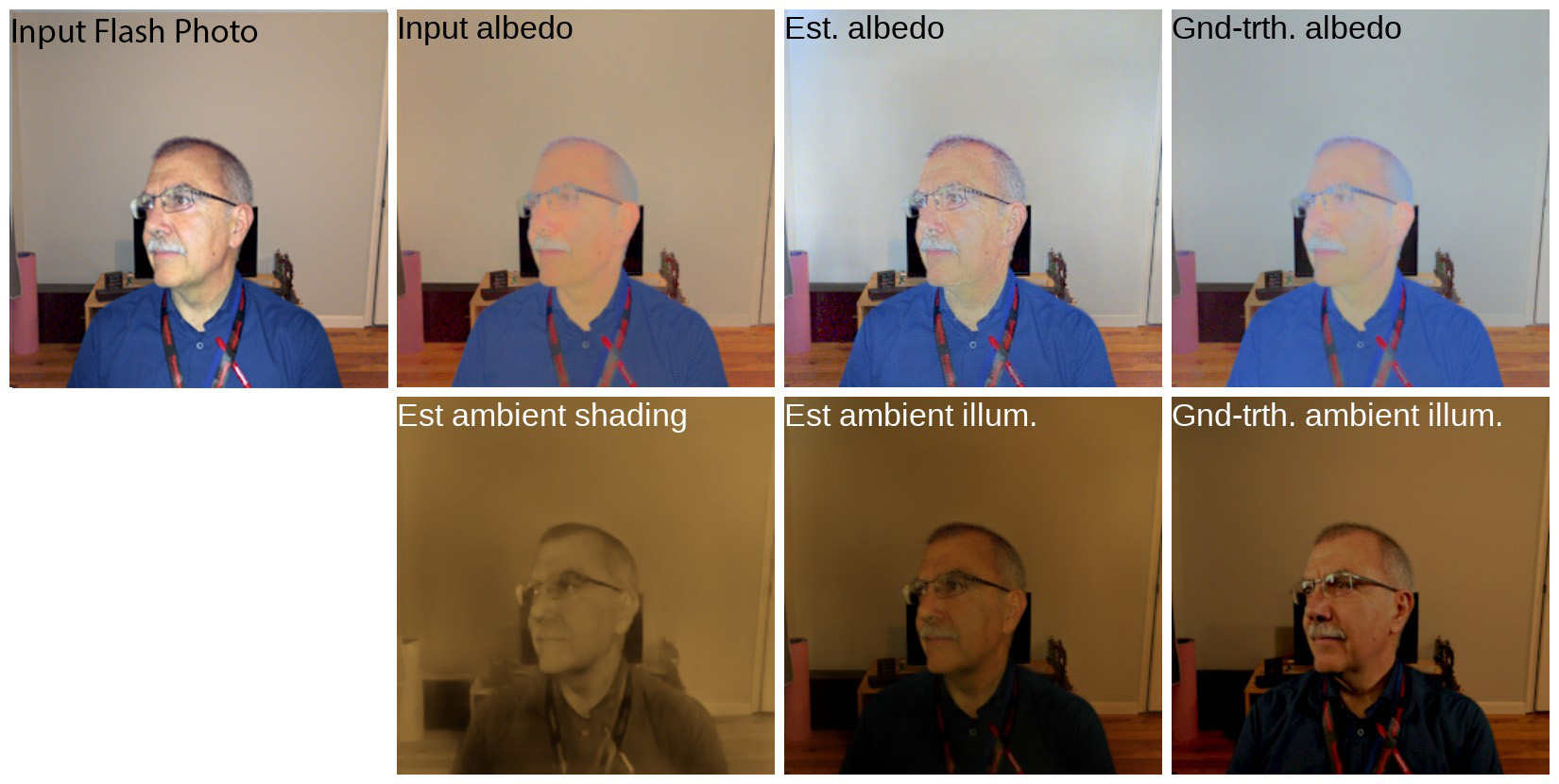}
\negvspace
\negvspace
\negvspace
\caption{We are able to disentangle the color of ambient illumination from the albedo of a flash photograph.
}
\negvspace
\negvspace
\label{fig:albedo_color}
\end{figure}

\vspace{0.3cm}
\noindent\textbf{Inputs}\quad
Since we define the decomposition problem in the shading domain, we also provide the intrinsic decomposition of the input flash photograph in terms of an albedo map. 
We generate this input using an off-the-shelf intrinsic decomposition method~\cite{chrisIntrinsic}. 
The input image and this albedo map together provide the information on the combined shading maps of ambient and flash, as these variables are related through Eq.~\ref{eq:intrinsicmodel}.
Since this method estimates a single-channel shading for the flash photograph, the color shift of the ambient illumination is baked into this albedo map, as Fig.~\ref{fig:albedo_color}. 
Since the amount of this color shift is directly related to the strength of the ambient illumination at a given pixel, the input albedo also provides important cues for the relative strength of ambient and flash shadings.

The flash light can be modeled as a point light source close to the camera center, with its direction parallel to the main camera axis. 
This makes the strength of the flash illumination at a given pixel highly correlated to the distance of that pixel to the camera following the inverse-square law of light. 
Similarly, it is also correlated with the surface normals in the scene which defines the incident angle of the flash light. 
We provide this geometric information to our network in the form of monocular depth~\cite{miangolehdepth} and surface normal~\cite{eftekhar2021omnidata} estimations using off-the-shelf methods.

\begin{figure*}[t]
\negvspace
\includegraphics[width=\linewidth]{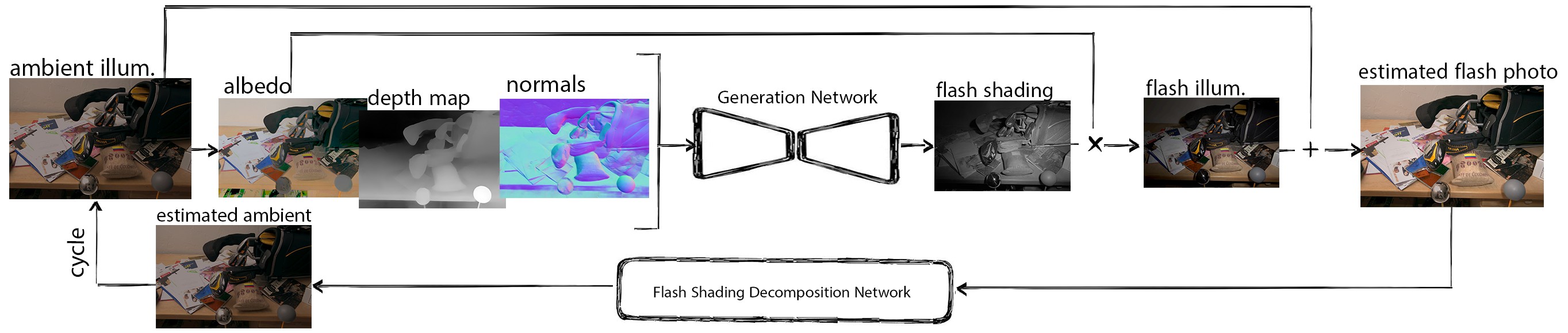}
\negvspace
\negvspace
\negvspace
\caption[Flash Shading Generation]{Albedo, normal, and depth maps are estimated from the ambient image and  provided to the encoder-decoder network. The network produces the flash shading, which is then used to generate the estimated flash photograph. Self-supervision is also done by cycle consistency with the pre-trained flash shading decomposition network.
}
\negvspace
\negvspace
\label{fig:generate}
\end{figure*}

\noindent\textbf{Outputs}\quad
The three decoders in our system generate estimates representing the scene illumination, $\hat{S}_A$, $\hat{S}_F$, and $\hat{c}_A$. 
Instead of estimating the 3-dimensional ambient illumination color $\hat{c}_A$ directly, we estimate the color temperature of the environment, $\hat{t}_a$, and compute $\hat{c}_a$ from it. 
This makes the color estimation problem single-dimensional while covering the plausible illumination conditions in natural scenes. 
Given these variables, using Eq.~\ref{eq:intrinsicsuperposition}, we can compute the common albedo:
\begin{equation}
    \hat{R} = \frac{P}{\hat{c}_A \hat{S}_A + \hat{S}_F}.
    \label{eq:impliedAlbedo}
\end{equation}
We then compute the individual flash-only and ambient-only illuminations using the intrinsic model:
\begin{equation}
    \hat{F} = \hat{R} \cdot \hat{S}_F, 
    \quad \quad \hat{A} = \hat{R} \cdot \hat{c}_A \hat{S}_A.
    \label{eq:decomposedillums}
\end{equation}

\noindent\textbf{Loss functions}\quad
We define two losses on the estimated shadings as well as the albedo derived from them in Eq.~\ref{eq:impliedAlbedo} for training our system. 
While the individual losses on shadings supervise each decoder separately, the loss on the albedo is backpropagated through all three decoders and ensures that the individual components combine well together to reconstruct the scene. 

The first loss we define is the standard $L_1$ loss:
\begin{equation}
    \mathcal{L}_{1,X} =  \frac{1}{N} \sum_{i=1}^N |\hat{X}_i - X^*_i|, \quad X \in \{S_A,S_F,R\},
    \label{eq:L1}
\end{equation}
where $X^*$ represents the ground truth. 
We also use the multi-scale gradient loss proposed by Li and Snavely~\cite{li2018megadepth} on these variables. 
This loss enforces spatial smoothness as well as sharp discontinuities guided by the ground truth:
\begin{equation}
    \mathcal{L}_{g,X} = \frac{1}{NM} \sum_{i=1}^N \sum_{j=1}^M |\nabla X_{i,l} - \nabla X^*_{i,l}|. 
    \label{eq:gradLoss}
\end{equation}
We use the standard $L_1$ loss on the color temperature, $\mathcal{L}_{T} = |\hat{t}_A - t^*_A|$, and combine all 7 losses to train our decomposition system:
%
%\begin{equation}
%    \mathcal{L}_{T} = |\hat{t}_A - t^*_A|,
%\end{equation}
%
%and combine all 7 losses to train our decomposition system:
%
\begin{equation}
    \begin{split}
        \mathcal{L}_D = & \mathcal{L}_{1,S_A} + \mathcal{L}_{1,S_F} + \mathcal{L}_{1,R} + \\
        & \quad\quad 0.5\left(\mathcal{L}_{g,S_A} + \mathcal{L}_{g,S_F} + \mathcal{L}_{g,R} \right) + \mathcal{L}_{T}.
    \end{split}
\end{equation}

\noindent\textbf{Network structure}\quad
We base our network on the encoder-decoder architecture proposed in~\cite{MiDas}. 
We use the EfficientNet~\cite{tan2019efficient} encoder and a pair of RefineNet~\cite{refinenet} decoders for flash and ambient shadings. The third decoder for the ambient color temperature uses global average pooling to flatten the activations from the encoder. 
The result is fed through 3 linear layers separated by ReLU activations. A final sigmoid activation outputs the color temperature. All components are trained from scratch with no pre-training with the learning rate of $2 \times 10^{-4}$ for 100 epochs.

\section{Flash Illumination Generation}
\label{sec:generation}

Similar to our decomposition formulation, we model our flash illumination generation method from a no-flash photograph in the intrinsic domain and estimate the single-channel flash shading. 
We supplement the supervised training with a self-supervised loss by taking advantage of the cyclic relationship between the generation and decomposition tasks.
Fig.~\ref{fig:generate} shows an overview of our pipeline.
%An overview of our generation pipeline is shown in Fig.~\ref{fig:generate}.

\vspace{0.3cm}
\noindent\textbf{Inputs}\quad
Similar to our decomposition setup, we estimate the albedo, the monocular depth, and the surface normals from the input no-flash photograph, which is the same as the ambient illumination. The flash illumination depends on the geometry and the albedo and is independent of the ambient shading as Eq.~\ref{eq:intrinsicsuperposition} shows. Once these variables are extracted, the original input does not provide additional useful information for the generation task. 
Hence, we exclude it from the inputs to our generation network.

\noindent\textbf{Outputs}\quad
Our generation architecture only estimates the flash shading $\hat{S}_F$. 
Assuming a no-flash photograph white balanced for the ambient illumination as input $A$, we can compute the flash illumination using the estimated shading and the input albedo and also generate a flash photograph for any ambient illumination color $c_A$:
\begin{equation}
    \hat{F} = R \cdot \hat{S}_F, \quad\quad \hat{P} = \hat{F} + c_A A.
    \label{eq:flashphotogen}
\end{equation}

\noindent\textbf{Self-Supervision through Cycle Consistency}\quad
Given the ambient illumination $A$, our network can generate a flash photograph $\hat{P}$ via Eq.~\ref{eq:flashphotogen}. 
Similarly, given a flash photograph $P$, the flash decomposition task can generate the ambient illumination $\hat{A}$ via Eq.~\ref{eq:decomposedillums}. 
This exposes a cyclic relationship between the two tasks, where the two networks cascaded should re-create the input image. 
We exploit this relationship to define an $L_1$ cyclic consistency loss~\cite{cyclegan} using a forward pass through decomposition network $\mathcal{D}(\cdot)$:
\begin{equation}
    \mathcal{L}_{cyc} = \frac{1}{N} \sum_{i=1}^N |\mathcal{D}(\hat{P})_i - A_i|.
    \label{eq:cycleloss}
\end{equation}
%
%where $\mathcal{D}(\cdot)$ represents a forward pass through the decomposition network.
%
When the decomposition network is at its initial stages of training, this loss is not very reliable given the unstable $\mathcal{D}(\hat{P})$ in Eq~\ref{eq:cycleloss}. 
Hence, we first train our decomposition task to completion before starting to train our generation network. 
Although this cyclic relationship can also be exploited in the other direction for the decomposition task, due to the higher complexity of the generation problem, we observed that it did not improve our decomposition performance.

\noindent\textbf{Losses and Network Structure}\quad
In addition to the cyclic consistency loss, we utilize $L_1$ and the multi-scale gradient losses defined in Eqs.~\ref{eq:L1} and \ref{eq:gradLoss} on the estimated shading $\hat{S_F}$ as well as the corresponding flash illumination $\hat{F}$:
\begin{equation}
    \mathcal{L}_G = \mathcal{L}_{1,S_F} + \mathcal{L}_{1,F} +
                    0.5 \left( \mathcal{L}_{g,S_F} + \mathcal{L}_{g,F} + \mathcal{L}_{cyc} \right).
\end{equation}  

We adopt the same network architecture as our decomposition task described in Sec.~\ref{sec:decomposition}, with the only difference being the use of a single decoder as only a single shading map is estimated in the generation task.

%\section{Generating High-Resolution Output}
%\label{sec:highres}
%\input{tex/5-highres}

\section{Dataset Preparation and Augmentation}
\label{sec:dataset}
\begin{figure}
\centering
\includegraphics[width=1\linewidth]{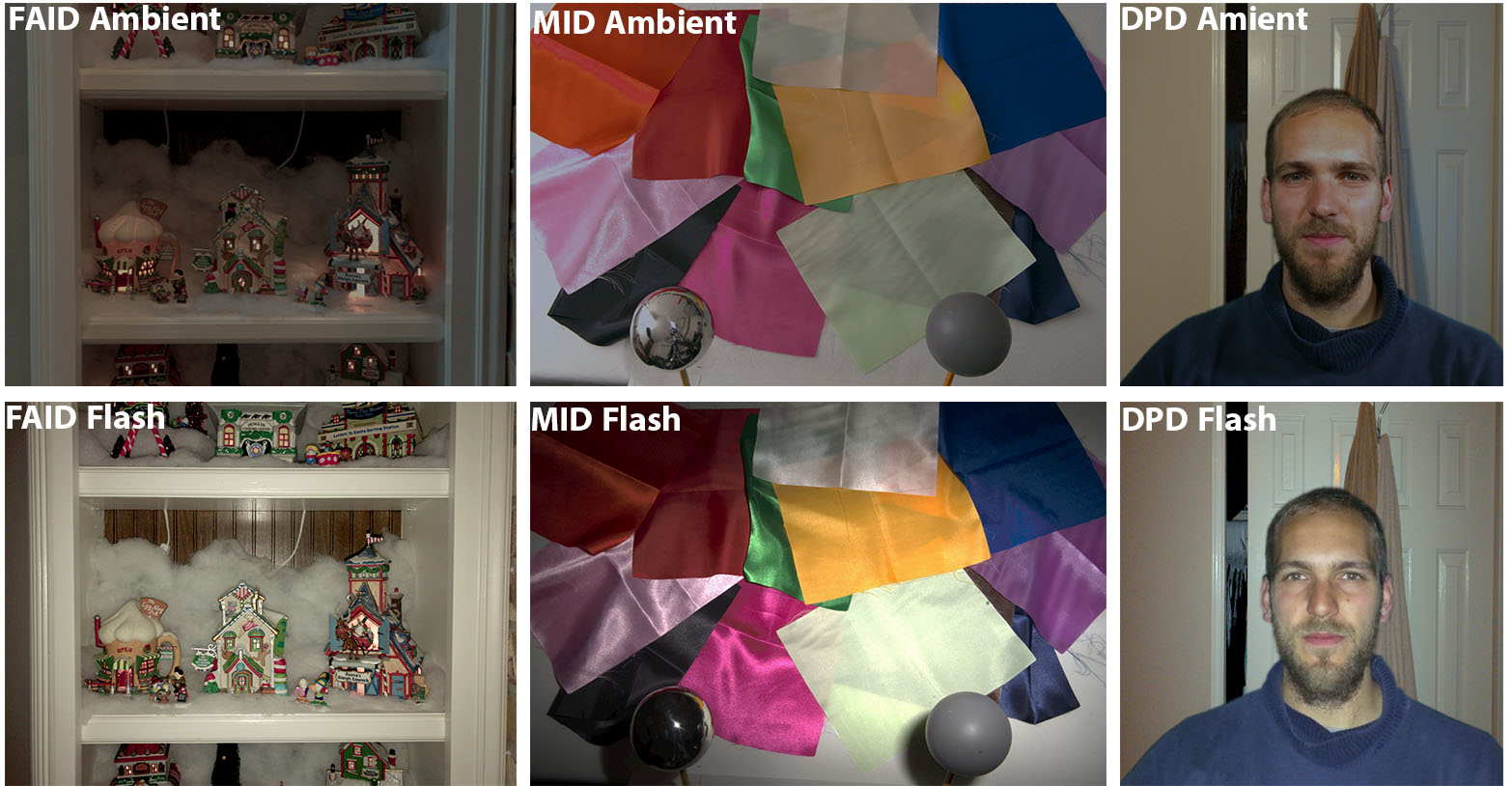}
\negvspace
\negvspace
\negvspace
\caption{Our dataset consists of flash/no-flash pairs from FAID~\cite{flashambient}, MID~\cite{multi2019} and DPD~\cite{deepflash} composited on background images from FAID. 
}
\negvspace
\negvspace
\label{fig:dataset}
\end{figure}

We require matching pairs of flash/no-flash pairs alongside their intrinsic decomposition components to train our networks. 
The flash decomposition and generation are challenging tasks that would benefit from a large training set. 
However, the availability of flash/ambient illumination pairs is limited due to the challenging capture process, and the number of input/ground-truth pairs is low when compared to datasets aiming at other relighting tasks such as facial relighting.
Hence, we combine the 3 available datasets for flash illumination, namely the Flash and Ambient Illuminations Dataset (FAID)~\cite{flashambient}, the Deep Flash Portrait Dataset (DPD)~\cite{deepflash} and the Multi-Illuminations Dataset (MID)~\cite{multi2019} and modify them to meet our needs. 
The FAID dataset contains pairs of flash/no-flash images, the DPD dataset includes portraits captured with and without flash in front of a green screen, and the MID dataset has 25 lighting directions for each scene. 
To generate realistic images from the green-screen data in DPD, we apply green-screen keying and add flash/no-flash background pairs to each illumination taken from the Rooms set in FAID. 
Examples from each dataset are shown in Fig.~\ref{fig:dataset}. 
The DPD is composited onto various room environments to create realistic-looking images, and in MID, the lighting from the front of the camera is selected as the flash illumination, while the other illuminations serve as the no-flash images. 
The brightness of flash and no-flash illuminations are normalized across the datasets. 
For the intrinsic components, we use the MID-Intrinsics dataset extension by \cite{chrisIntrinsic} for MID and use the robust multi-illumination intrinsic decomposition approach~\cite{chrisIntrinsic} to generate the albedo and shading pairs for FAID and DPD. 
We present a detailed description of our dataset pre-processing pipeline in the supplementary material.

\section{Experimental Analysis}
\label{sec:experimental}

\begin{figure} \centering
\includegraphics[width=1\linewidth]{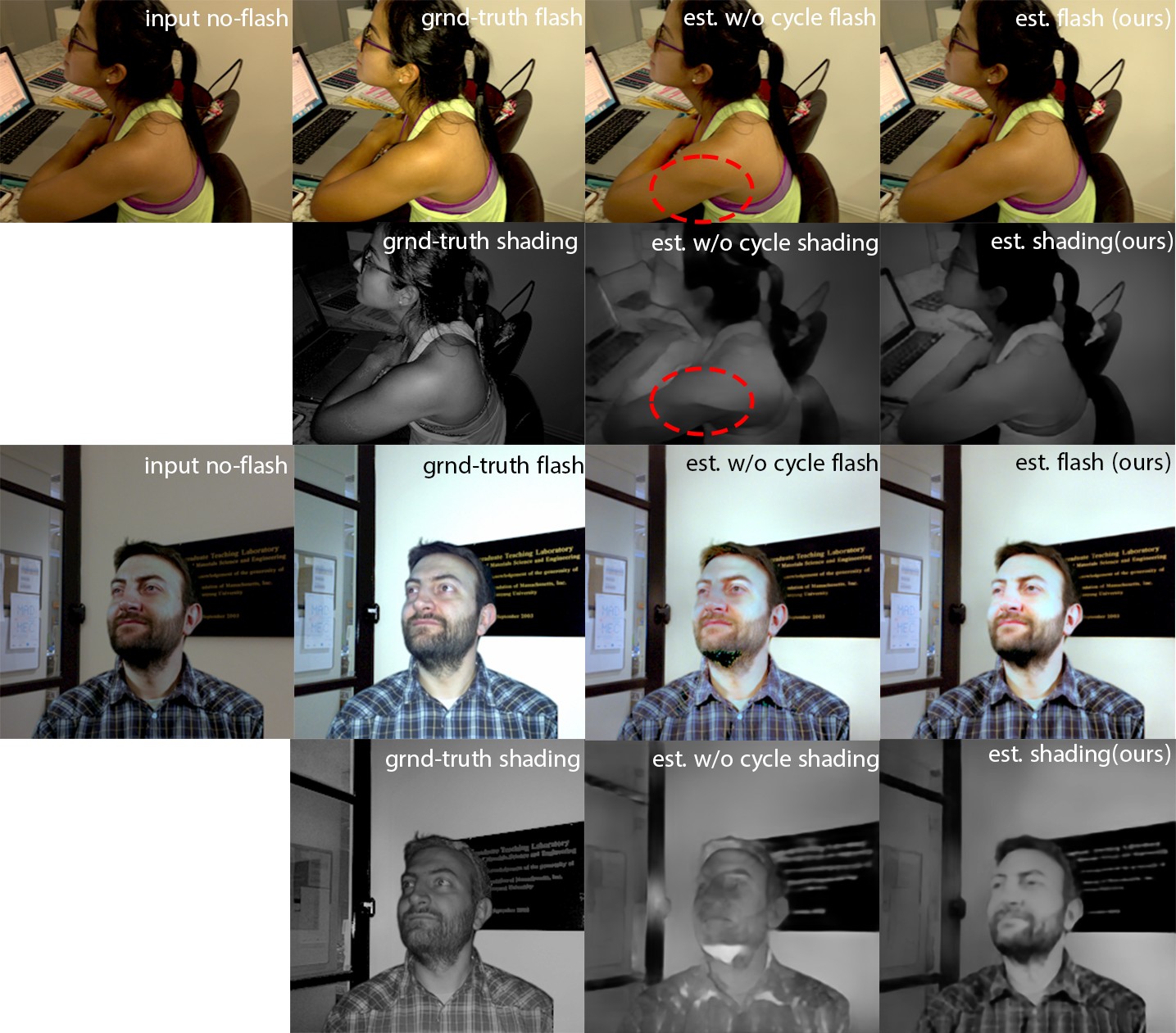}
\negvspace
\negvspace
\negvspace
\caption[Cycle Consistency Effect on Shading Estimation]{
We utilize the cycle consistency loss for generation. The effect of this loss is evident in the shadings depicted by the red circle in the first example, as well as on the man's face in the bottom.% These precise estimations could also eliminate incorrect illuminations in the final image.
}
\negvspace
\negvspace
\label{fig:abl_cycle_sh}
\end{figure}
\begin{figure} \centering
\includegraphics[width=1\linewidth]{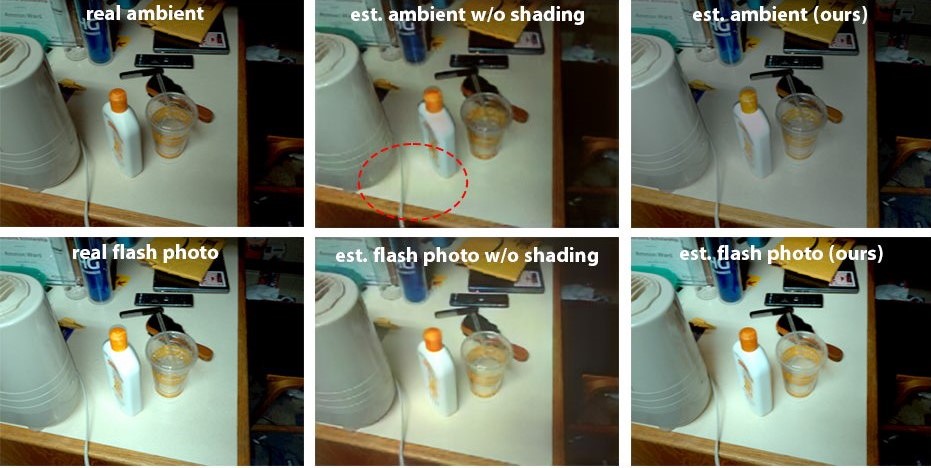}
\negvspace
\negvspace
\negvspace
\caption[Difference between shading estimation and illumination estimation]{By estimating the shadings instead of estimating the illumination, the decomposition is incomplete in certain areas shown in red circles and the visual quality of estimation degrades.}
\negvspace
\negvspace
\label{fig:simple}
\end{figure}

To show the effectiveness of our proposed approach, we perform an ablation over the various design choices of our method, as well as comparisons to various prior works. We utilize three common metrics for evaluating performance on both flash decomposition and generation. Namely, structural similarity index measure (SSIM), peak signal-to-noise ratio (PSNR), and Inception Score (IS)~\cite{IS}. PSNR uses mean-square error to do a pixel-by-pixel comparison between the estimation and ground truth, while SSIM and IS focus on image quality. SSIM utilizes structure, luminance, and contrast to estimate perceived similarity to the ground truth. IS measures high-level image generation quality using an Inception-v3 image classifier~\cite{Inception} network trained on ImageNet~\cite{imagenet}.

\begin{figure}
\centering
\includegraphics[width=1\linewidth]{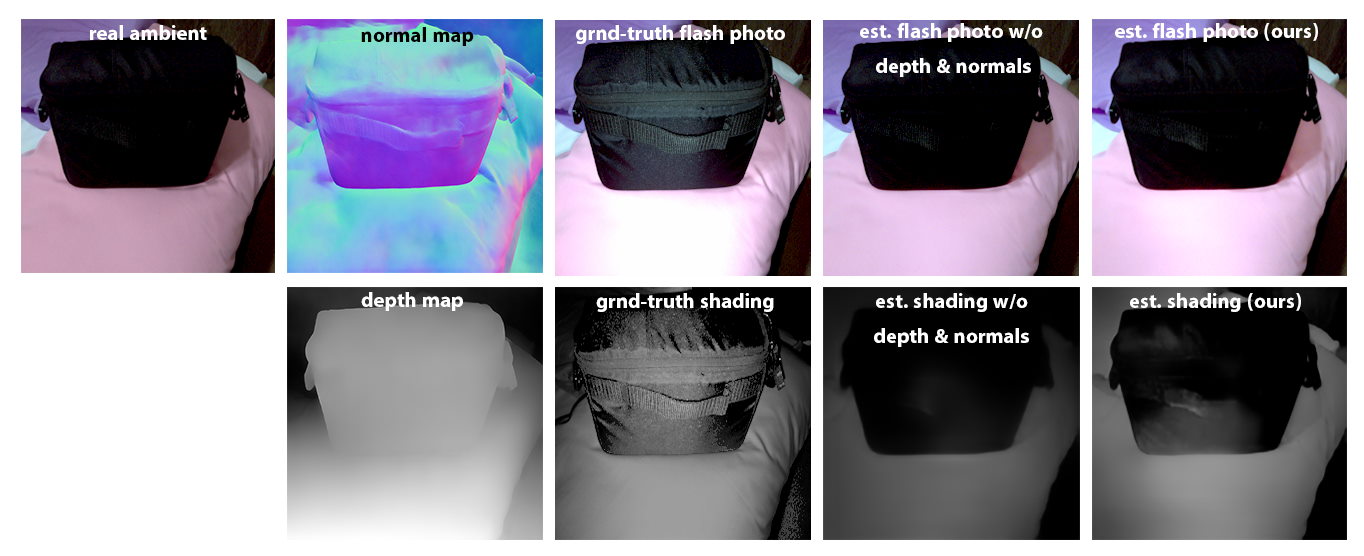}
\negvspace
\negvspace
\negvspace
\caption[Effect of monocular depth estimation and surface normals to the shading estimation]{
The monocular depth and surface normals provide geometric details that contribute to more accurate flash illumination. 
}
\negvspace
\negvspace
\label{fig:depth_abl_sh}
\end{figure}
\begin{figure}\centering
\includegraphics[width=1\linewidth]{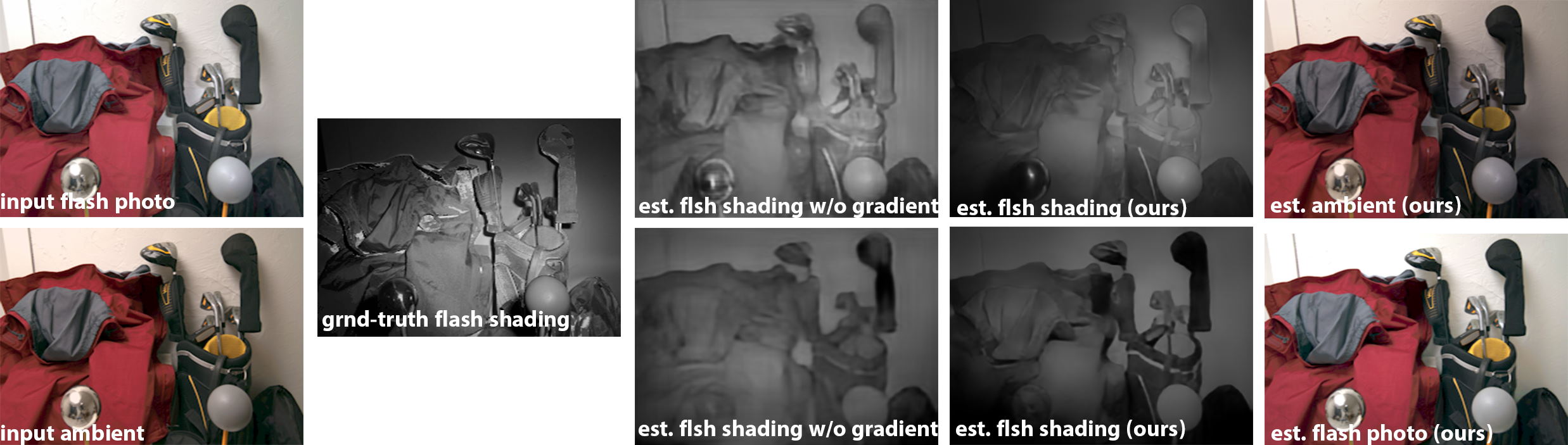}
\negvspace
\negvspace
\caption[Effect of multi-scale gradient term on shading estimations]{
We utilize the multi-scale gradient loss to create sharper edges and smoother gradients in the flash shading estimations. %This term is useful for both generation and decomposition.
}
\negvspace
\negvspace
\label{fig:grad_abl}
\end{figure}

\subsection{Ablation Study}
The top half of Table~\ref{tab:shading_abl} shows the quantitative results of our ablation tests for 200 images containing a variety of objects, scenes, and people never seen before by our methods. Each row shows our pipeline with a single aspect missing. 

%\textbf{Intrinsic Formulation}\quad
We first examine the efficacy of our proposed intrinsic formulation. We train a version of our pipeline that directly regresses the final relit image for both the generation and decomposition tasks. This causes a significant drop in quantitative performance across all metrics. Without our intrinsic formulation, the network has to reproduce redundant RGB information that is present in the input image, greatly increasing the likelihood of artifacts. Our proposed approach on the other hand disentangles the contributions of reflectance and shading and the network, therefore, has a much easier job of estimating single-channel shading images. Fig.~\ref{fig:simple} shows some examples from our model trained with and without the intrinsic formulation.

%\textbf{Cycle Loss}\quad
For the generation task, we evaluate our approach without the proposed cycle loss. We observe that while the low-level metrics, PSNR and SSIM, remain the same, the cycle loss significantly degrades the perceptual quality of our generated flash as measured by the Inception Score. Since the cycle loss acts as a complementary form of supervision for the model, we can generate a more plausible flash. Some example outputs from our generation model with and without the cycle loss are shown in Fig.~\ref{fig:abl_cycle_sh}. 

%\textbf{Multi-Scale Gradient Loss}\quad
The multi-scale gradient loss allows us to generate more accurate shading maps at both sharp discontinuities and over smooth image regions. In Fig.~\ref{fig:grad_abl} we show some examples of our shading estimations, with and without the multi-scale gradient loss for both generation and decomposition.

\begin{table}[]
\caption{Quantitative performance of different ablation configurations and alternative methods on PSNR, SSIM and IS~\cite{IS}. %$\uparrow$ indicates higher is better.
}
\negvspace
\begin{adjustbox}{width=\linewidth}
{\renewcommand{\arraystretch}{1.5}}
\begin{tabular}{l|lll|lll}
\hline
\multicolumn{1}{c|}{\multirow{2}{*}{Method}} & \multicolumn{3}{c|}{Decomposition} & \multicolumn{3}{c}{Generation} \\
\multicolumn{1}{c|}{}                           & SSIM$\uparrow$    & PSNR$\uparrow$   & IS\cite{IS}$\uparrow$   & SSIM$\uparrow$   & PSNR$\uparrow$   & IS\cite{IS}$\uparrow$ \\ \hline
Ours                                         & \textbf{0.8553} & 19.6044  & \textbf{7.2892} &  \textbf{0.8601} & 19.1863 &  \textbf{7.3730} \\
no albedo/shading
 & 0.774 & 19.2296 & 6.0548  & 0.7352 & 14.5329 &  5.9953 \\
no cycle loss                            & - & - & - & 0.8597 & 19.2155 & 6.4777 \\
no gradient loss                            &  0.8096 & 16.2628  &  6.4617 & 0.8573 & 19.0036 &  6.5127      \\
no geometry                                    & 0.8074 & 14.5056 & 6.4746  & 0.8584 & 19.1301 & 6.5307   \\
\hline
DeepFLASH  \cite{deepflash}                   & 0.5744 & 13.2684 & 7.1276  & - & - & - \\ %0.7548 & 14.4607 &  \textbf{7.6142}     \\
FAID network\cite{flashambient}             & 0.8081 & 15.6040 &  7.0183 & 0.7830 & 14.1004 & 6.7445      \\
Pix2pixHD \cite{pix2pixHD}                   & 0.7917 & 17.486 & 6.5667 & 0.6866 & 12.2330 &  5.74385     \\ 
IAN \cite{zhu2022ian}                        & {0.8507} & \textbf{21.4149} &   7.0240 & \textbf{0.8682} & 19.9776 & {7.2694}      \\ 
OIDDR \cite{Yazdani2021PhysicallyID}         & {0.8547} & 20.8643 &  {7.2035} &  0.8579 & \textbf{20.4645} &  7.1679   \\ 
\hline
\end{tabular}
\end{adjustbox}
\negvspace
\negvspace
\label{tab:shading_abl}
\end{table}

\begin{figure*}
\includegraphics[width=1\linewidth]{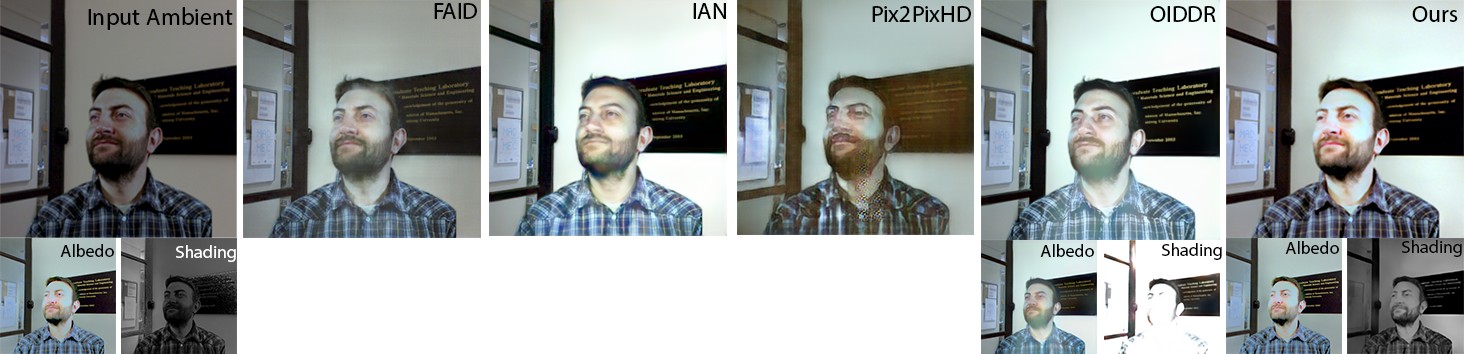}
\negvspace
\negvspace
\caption{Comparison of generation task with different methods. We also show the generated albedo and shading from the OIDDR~\cite{Yazdani2021PhysicallyID} which estimations are also done through intrinsics.
Please refer to text for detailed discussions.
}
\negvspace
\label{fig:comparisonGeneration}
\end{figure*}

\begin{figure*}
\includegraphics[width=1\linewidth]{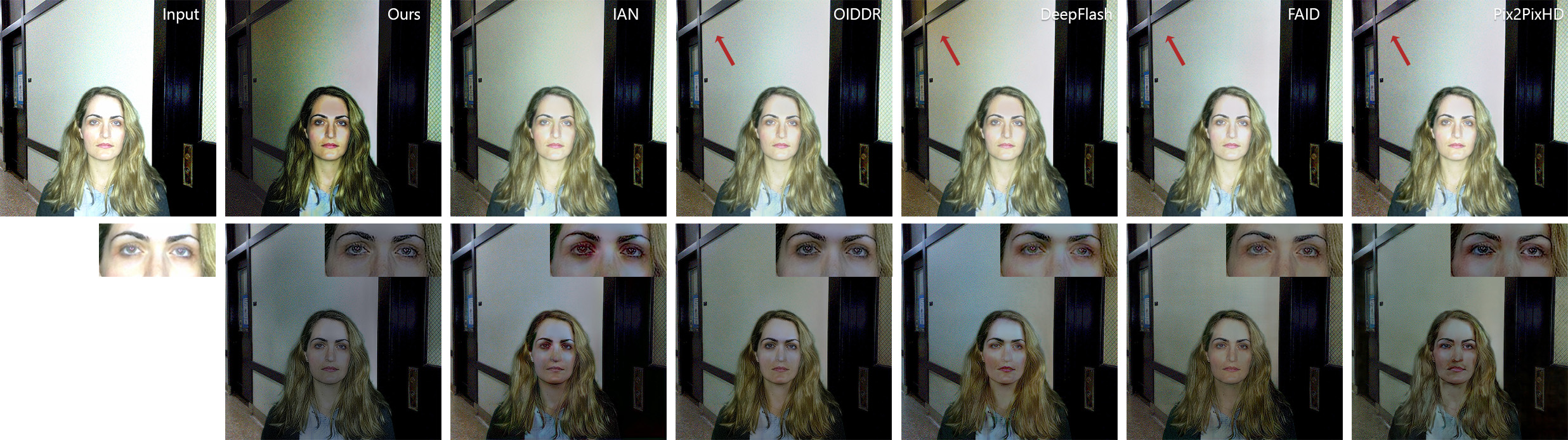}
\negvspace
\negvspace
\caption{Comparison of decomposition task with different methods. Please refer to text for detailed discussions.
}
\negvspace
\negvspace
\label{fig:comparisonDecomposition}
\end{figure*}

%\textbf{Geometry Information}\quad
Naturally, geometry information is very informative for relighting tasks as lighting effects are directly correlated with geometry. We find that removing the depth and surface normal input from our formulation greatly affects performance in both generation and decomposition. Fig.~\ref{fig:depth_abl_sh} shows an example of our flash generation task with and without geometry information.

\subsection{Comparison to Prior Work}
Although no prior works attempt to generate flash illumination in such diverse input images, to thoroughly evaluate our proposed pipeline, we use our augmented dataset to train five networks from prior works for our generation and decomposition tasks. We train all the networks for 100 epochs and compare them qualitatively using the previously discussed metrics. The bottom half of Table~\ref{tab:shading_abl} shows the quantitative comparison of these methods on the same test set as the ablation study.

%\paragraph{FAID~\cite{flashambient} and Pix2PixHD~\cite{pix2pixHD}}
% \yagiz{They are very similar, both image-to-image translation, Pix2PixHD is just an improved network over FAID}
We compare our approach to two image-to-image baseline networks. The baseline network utilized by Aksoy et al.~\cite{flashambient} and Pix2PixHD~\cite{pix2pixHD} are both based on the original Pix2Pix UNet architecture~\cite{unet}. These baselines exhibit severe artifacts, exemplifying the inherent difficulty of learning these tasks using a relatively small-scale dataset.

% \paragraph{DeepFLASH~\cite{deepflash}}
We follow the training scheme proposed by Capece et al.~\cite{deepflash} and train their proposed model on our augmented dataset. To train the network, we had to adjust the beta parameters of the Adam optimizer, specifically, we set $\beta_1$ equal to 0.5. 
We failed to train this method for the generation task, in which it simply outputs the input without modification, and hence we exclude it from our evaluation. On the decomposition task, while their method obtains a competitive Inception Score, they perform poorly on low-level metrics measured against the ground truth. This indicates that while their outputs may be visually coherent, they are not able to separate flash and ambient lighting effects.

% \paragraph{IAN~\cite{zhu2022ian}}
Zhu et al.~\cite{zhu2022ian} propose a direct estimation approach for relighting between two known lighting configurations. They propose multiple architectural improvements specifically designed to achieve better relighting. We find that due to their carefully designed network, they can achieve competitive quantitative results on the test set. However, given their direct approach to relighting, their model still produces noticeable artifacts on difficult scenes such as portrait photographs, as can be seen in the insets of Fig.~\ref{fig:comparisonDecomposition}.

% \paragraph{OIDDR~\cite{Yazdani2021PhysicallyID}}
Similar to our method, the work of Yazdani et al.~\cite{Yazdani2021PhysicallyID} utilizes intrinsic components as part of their proposed pipeline. We are therefore able to supervise their model using our generated shading and albedo components. Unlike our work, they propose to estimate both the shading and albedo of the relit output. Given that intrinsic decomposition is a notoriously difficult problem, this adds extra complexity to the relighting problem. Our method, on the other hand, utilizes albedo estimations as input, and in the case of the flash generation directly makes use of the albedo to compute the flash illumination. By only estimating grayscale shading, we significantly reduce the complexity of the task and generate stable, artifact-free estimations. Additionally, we can directly compare our predicted intrinsic components to theirs. In the generation task, we observe that their predicted albedo looks very similar to their final output, as seen in Fig.~\ref{fig:comparisonGeneration}. Due to the sigmoid activation they use for shading, their approach fails to properly represent the high magnitude flash shading, causing the flash illumination to leak into the albedo. This indicates that their model is likely not benefiting from the intrinsic modeling to the full extent. 

While the baseline approaches are not able to compete with our method quantitatively, the more recent methods IAN~\cite{zhu2022ian} and OIDDR~\cite{Yazdani2021PhysicallyID} obtain competitive quantitative scores. There are multiple factors contributing to the discrepancy between quantitative scores and qualitative observations. Firstly, although our physically-inspired formulation generates more stable outputs, our modeling comes with some constraints. For example, in the generation task, our model relies on the input albedo and grayscale shading to generate the flash illumination. Oftentimes the mapping from ambient albedo to flash illumination cannot be expressed as a single per-pixel multiplier. This means that while our results are without artifacts, we may not be able to estimate the exact appearance of the ground-truth flash photograph. In the decomposition task, we are estimating the color temperature of the ambient illumination. While this design choice simplifies the shading estimation, a small deviation from the ground-truth color temperature can result in errors across the whole image. Despite these factors, our model still achieves scores that are higher or on par with recent relighting works in terms of SSIM and PSNR. Furthermore, our model outperforms all other competing methods in terms of Inception Score which correlates with the visual quality estimations. Given that we achieve these scores using off-the-shelf networks with no architectural improvements, we attribute our performance to our careful modeling of flash photography.

\begin{figure*}
\includegraphics[width=\linewidth]{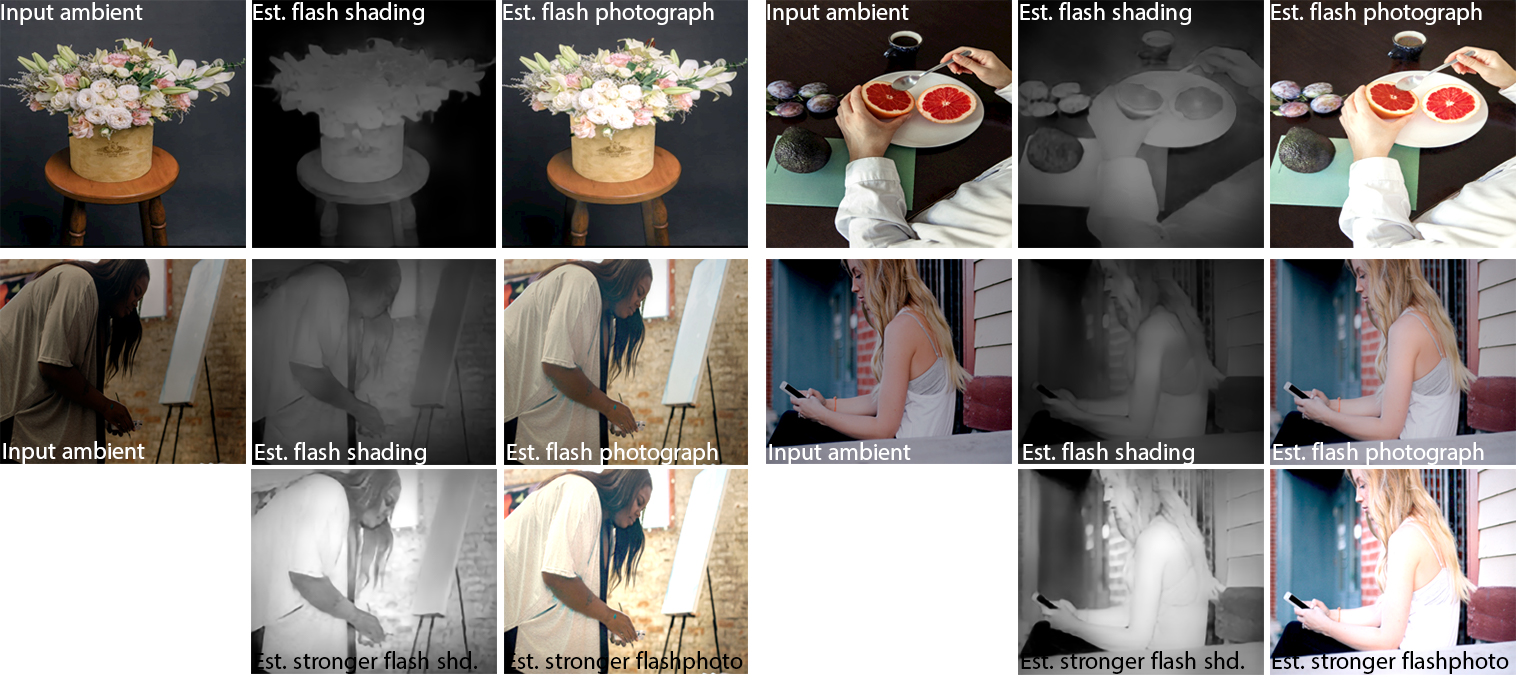}
\negvspace
\negvspace
\caption[Generation of flash shading]{
We show examples of our flash generation method on in-the-wild photographs with controllable flash strength.
}
\negvspace
\label{qual_gen}
\end{figure*}
\begin{figure*}\centering
\includegraphics[width=\linewidth]{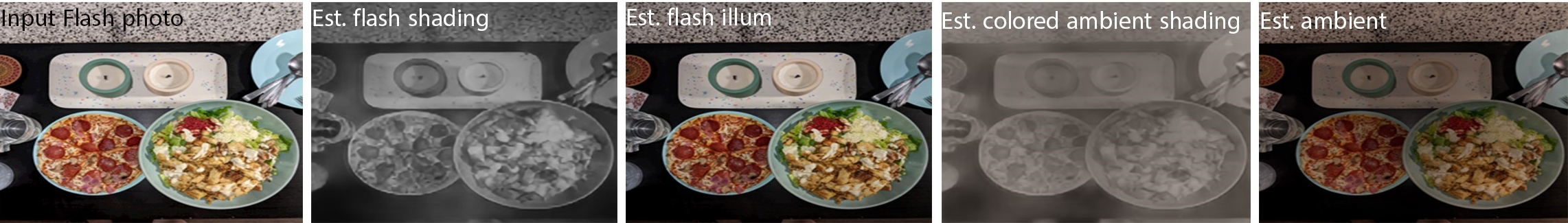}
\negvspace
\negvspace
\caption[Decomposition flash and ambient shadings]{
Our intrinsic shading decomposition is able to decompose a flash photograph into its flash and ambient illuminations. 
}
\negvspace
\negvspace
\label{fig:qual_dec}
\end{figure*}
\begin{figure}\centering
\includegraphics[width=\linewidth]{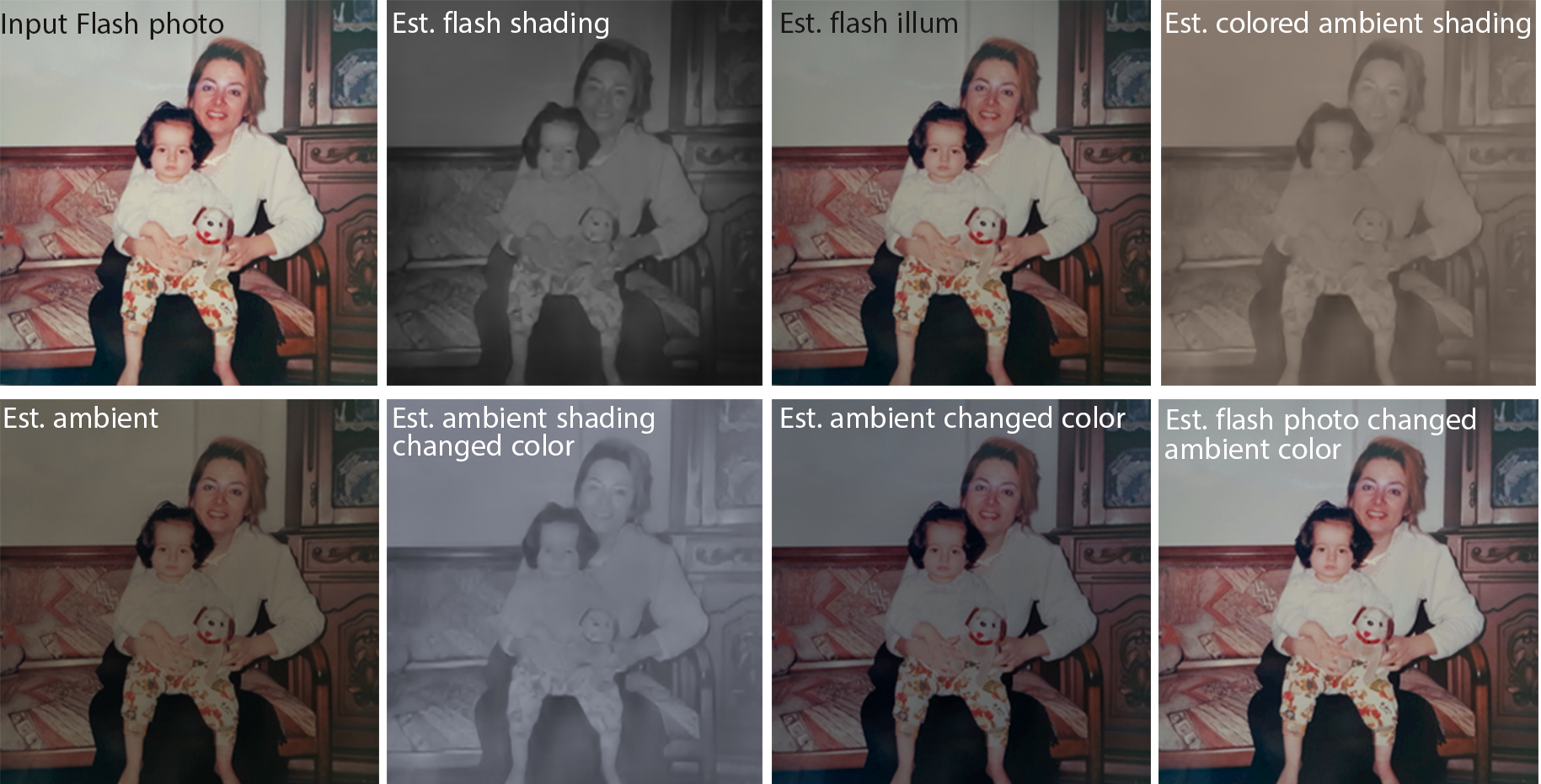}
\negvspace
\negvspace
\caption[Changing color of estimated ambient shading]{
The decomposed ambient and flash illuminations allow us to control the ambient illumination color independent of the flash light.
}
\negvspace
\negvspace
\label{fig:qual_dec_comp}
\end{figure}
\subsection{Computational Flash Photography}
Our decomposition and generation results can be used to edit or simulate the flash light for flash or no-flash photographs. 

We show in-the-wild examples with simulated flash in Fig.~\ref{qual_gen}. 
Fig.~\ref{fig:qual_dec} shows our decomposition with the estimated shadings, while Fig.~\ref{fig:qual_dec_comp} shows an illumination editing example on an image originally captured on film.

% \section{Computational Flash Photography}
% \label{sec:applications}
% \input{tex/8-applications}

\section{Limitations}
\label{sec:limitation}

Flash generation and decomposition are challenging image relighting problems that require a large amount of training data. 
Even after combining 3 datasets available for our tasks, the volume of our dataset measures in a few thousand unique flash-ambient pairs. 
Similar tasks such as portrait relighting are typically achieved using datasets that are a magnitude larger, measuring in tens of thousands. 
This makes it hard to model challenging areas such as specularities.
Still, our intrinsic formulation helps us generate high-quality results when compared to image-to-image translation under this limited training data. 
Our generation model takes the albedo, monocular depth, and surface normals as input. 
While, physically, this is the complete data necessary for flash generation, errors in the estimations of these variables may be carried to our final estimations.

\negvspace
\section{Conclusion}
\label{sec:conclusion}
In this work, we presented an intrinsic model for flash photograph formation and applied it to two tasks in computational flash photography: illumination decomposition and flash light generation. 
We show that, despite the limited training data, our physically-inspired formulation allows us to generate realistic results and control the flash light computationally in everyday photography. 
Our experimental evaluation demonstrates that, compared to image-to-image translation methods commonly utilized for similar tasks in the literature, our intrinsic formulation achieves higher-quality results.

\paragraph{Acknowledgements}
We would like to thank Sebastian Dille and S. Mahdi H. Miangoleh for our discussions and their support with the dataset and training setups.

%%%%%%%%% REFERENCES
{\small
\bibliographystyle{ieee_fullname}
\bibliography{main}
}

\end{document}